\documentclass[10pt, conference, compsocconf]{IEEEtran}
\ifCLASSINFOpdf
\else
\fi

\usepackage{graphicx}
\usepackage{subfigure}
\usepackage{adjustbox}
\usepackage{amsmath}
\DeclareMathOperator*{\argmax}{argmax}

\usepackage{algorithm,algorithmic}

\usepackage{soul}
\begin{document}
\title{Multi-objective Reinforcement Learning based approach for User-Centric Power Optimization in Smart Home Environments}


% \author{Anonymous Submission}

% \author{\IEEEauthorblockN{Authors Name/s per 1st Affiliation (Author)}
% \IEEEauthorblockA{line 1 (of Affiliation): dept. name of organization\\
% line 2: name of organization, acronyms acceptable\\
% line 3: City, Country\\
% line 4: Email: name@xyz.com}
% \and
% \IEEEauthorblockN{Authors Name/s per 2nd Affiliation (Author)}
% \IEEEauthorblockA{line 1 (of Affiliation): dept. name of organization\\
% line 2: name of organization, acronyms acceptable\\
% line 3: City, Country\\
% line 4: Email: name@xyz.com}
% }



\author{\IEEEauthorblockN{Saurabh Gupta\IEEEauthorrefmark{1},
Siddhant Bhambri\IEEEauthorrefmark{2},
Karan Dhingra\IEEEauthorrefmark{1}, 
Arun Balaji Buduru\IEEEauthorrefmark{1} and
Ponnurangam Kumaraguru\IEEEauthorrefmark{1}}
\IEEEauthorblockA{\IEEEauthorrefmark{1}Indraprastha Institute of Information Technology,
Delhi, India\\ \{saurabhg,karan19025,arunb,pk\}@iiitd.ac.in }
\IEEEauthorblockA{\IEEEauthorrefmark{2}Delhi Technological University, Delhi, India\\
siddhant\_bt2k16@dtu.ac.in}
}


\maketitle

\begin{abstract}
Smart homes require every device inside them to be connected with each other at all times, which leads to a lot of power wastage on a daily basis. As the devices inside a smart home increase, it becomes difficult for the user to control or operate every individual device optimally. Therefore, users generally rely on power management systems for such optimization but often are not satisfied with the results. In this paper, we present a novel multi-objective reinforcement learning framework with two-fold objectives of minimizing power consumption and maximizing user satisfaction. The framework explores the trade-off between the two objectives and converges to a better power management policy when both objectives are considered while finding an optimal policy. We experiment on real-world smart home data, and show that the multi-objective approaches: i) establish trade-off between the two objectives, ii) achieve better combined user satisfaction and power consumption than single-objective approaches. We also show that the devices that are used regularly and have several fluctuations in device modes at regular intervals should be targeted for optimization, and the experiments on data from other smart homes fetch similar results, hence ensuring transfer-ability of the proposed framework. 
\end{abstract}


% \fi
\IEEEpeerreviewmaketitle

\section{Introduction}

\noindent The amount of power consumption in households (residential) is among the top three in world electricity consumption \cite{iea}, and is ever increasing with the increase in demand of smart homes and IoT (Internet of Things) devices. According to the United States Department of Energy (DoE), the average household consumes 90 million units of power a year, and much of that power is wasted \cite{con}. Habits like leaving lights on when we leave rooms, forgetting to turn off televisions or computers when not in use, etc., are major reasons behind such wastage \cite{cons}. Therefore, there is a need for power controllers that can take actions like turning devices on and off, or changing devices' modes of operation on behalf of users to achieve a goal like optimized consumption.

In the past, researchers have used traditional reinforcement learning for several power optimization tasks. For example, \cite{tan2009adaptive} proposed a model-free constrained RL approach for online power management. \cite{shen2013achieving} presented another similar algorithm that requires no prior information of the workload and dynamically adapts to the environment to achieve autonomous power management. \cite{tesauro2008managing} proposed an RL based technique that performs simultaneous online management of both performance and power consumption. The authors applied RL in a realistic laboratory testbed to find the optimal policy. None of these techniques towards power optimization are used for smart home power management, and they do not consider user satisfaction while finding optimal policies. 

However, power management in a smart home is a problem that needs to solve two tasks with different rewards simultaneously: minimize power consumption and maximize user satisfaction. It is important for a power controller to consider user preferences as well, i.e., the goal of minimal power consumption must be achieved but not at the expense of user satisfaction. The scenario can be formulated as a MORL problem where sequential decision making is required with multiple objectives.

 \textit{Our contribution:} In this paper, we propose, for the first time, a novel multi-objective reinforcement learning (MORL) approach for power management inside a smart home with two objectives: minimize power consumption and maximize user satisfaction. In a MORL problem, an action on the environment results in multiple rewards. The agent (power controller) learns an optimal policy from these rewards using a variation of Q-learning \cite{watkins1992q}. Since the objectives are contrasting, there is a trade-off between the two, and based on their importance, optimization priorities are set. We use an overall reward function to incorporate these optimization priorities, which is a weighted sum of the two rewards $R_E$ representing power consumption, and $R_U$ representing user satisfaction. We specifically focus on the weighted-sum method \cite{kim2006adaptive} for multi-objective optimization and compare the results with single objective strategies.

We evaluate our proposed methods on the Smart* data set for sustainability\cite{Barker2012Smart}. The data samples include device-level real-world power consumption values in several smart homes, named as A, B, C, ..., H recorded at every 30 minutes. We show the effectiveness of our approach on data from smart home A, and demonstrate transferability of experiments on smart homes B and C. We use Q learning with individual objectives (single policy single objective approaches) as a baseline reference to compare the proposed single policy multi-objective approaches. We also define a metric ``clash rate'' for evaluating user satisfaction in the predicted policy at each episode.

The remainder of this paper is organized as follows. Section \ref{section:background} gives some background of traditional and multi-objective reinforcement learning (MORL). Section \ref{section:problem} explains our problem formulation followed by our algorithm to solve the optimization problem in Section \ref{section:sol}. The experiments and results are presented in Section \ref{section:exp}-\ref{section:newres}. We conclude our work in Section \ref{section:conclude}.

\section{Background}
\label{section:background}
\noindent In this section, we discuss traditional reinforcement learning with Q-learning, an algorithm widely used to solve traditional RL problems. Then we introduce the concepts of multi-objective reinforcement learning (MORL) and how it differs from the traditional RL. 

\subsection{Traditional Reinforcement Learning}
\noindent Traditional reinforcement learning \cite{sutton2018reinforcement} mimics the natural learning style of trial-and-error by interacting with an environment (static or dynamic) and receiving feedback based on an action. The components of reinforcement learning are:

\begin{itemize}
    \item An Agent;
    \item A finite state space $S$;
    \item A set of available actions for the agent $A$;
    \item A reward function $R: S \times A \rightarrow R$.
\end{itemize}

 The agent's objective is to maximize its average long-term reward. It is achieved by learning a policy $\pi$, which is a mapping between the states and the actions. In our problem, one goal is to minimize the power consumption of a smart home, and the other is to maximize user satisfaction. But, in a traditional reinforcement learning setting, the two goals are independent. An agent can either minimize power consumption, or it can maximize user satisfaction. 

 Q-learning \cite{watkins1992q} is a widely known algorithm used to solve sequential decision-making RL problems. In each step, on the successful execution of every action $a$, the environment yields a reward $R$, which indicates the value of a state transition. The issued reward can be positive or negative. The agent keeps a value function $Q^{\pi}(s,a)$ for each state-action pair. Learning to act in the environment will make the agent choose actions to maximize long-term rewards. Based on this value function, the agent decides its immediate action. The Q-value for each state-action pair is initially chosen during the problem formulation, and later, it is updated with each taken action and its issued reward. The value function is given by the following Bellman equation:

\begin{equation}
\label{eq:q}
    Q^{\pi}(s, a) = R(s,a) + \gamma \max_a Q^{\pi}(s',a) 
\end{equation}

where $R(s,a)$ is the reward issued after taking action $a$ in state $s$, $s'$ is the successive state of $s$, and $\gamma$ is the discount factor used due to the different influences of future rewards on the present value.  

The optimal state-action value function is defined as:
\begin{equation}
    Q^{*} (s,a) = max_{\pi} Q^{\pi}(s,a)
\end{equation}

When $Q^{*} (s,a)$ is obtained, the optimal policy $\pi^*$ can be computed by:

\begin{displaymath}
\pi^*(s) = \argmax_a Q^*(s,a)
\end{displaymath}

% The complete Q-learning algorithm is shown in Algorithm \ref{alg:q}, where $\alpha$, the learning rate, and $\gamma$, the discount factor
% are the two hyper parameters to be tuned to get the best performance.

% \begin{algorithm}
%  \caption{Q-Learning Algorithm \cite{sutton2018reinforcement},\cite{watkins1992q}}
%  \begin{algorithmic}[1]
%  \label{alg:q}

% %  \renewcommand{\algorithmicrequire}{\textbf{Input:}}
% %  \renewcommand{\algorithmicensure}{\textbf{Output:}}
% %  \REQUIRE in
% %  \ENSURE  out
% %  \textit{Initialisation} :
%   \STATE N: the maximum number of episodes 
%   \STATE Initialize $Q(s,a)$ arbitrarily;
%   \FOR {each episode $i$ ranging from $1$ to $N$}
%   \STATE Initialize s;
%     \FOR{each step of episode}
%         \WHILE{$s$ is non-terminal}
%             \STATE Choose $a$ from $s$ using policy derived  from $Q(s,a)$;
%             \STATE Take action $a$, observe $r$, $s'$;
%             \STATE $Q(s,a) \leftarrow Q(s,a) + \alpha[r + \gamma \max_a' Q(s',a') - Q(s,a) ]$; 
%             \STATE $s \leftarrow s'$;
%         \ENDWHILE
% %   \IF {($i \ne 0$)}
% %   \STATE statement..
% %   \ENDIF
%     \ENDFOR
%   \ENDFOR
%  \end{algorithmic} 
%  \end{algorithm}

\subsection{Multi-objective Reinforcement Learning}
 \noindent Reinforcement learning is a machine learning paradigm that helps with sequential decision making under several uncertainties and aims to achieve a single long-term objective. However, due to the complex requirements of real-world control systems, often times, there are two (or more) conflicting objectives. For example, in our case of smart home power management system the controller has two goals: i) to minimize energy consumption of the smart home, ii) to maximize user comfort by moving to states preferred by the user. In reinforcement learning, problems of this nature having more than one conflicting objectives are called multi-objective reinforcement learning problems (MORL).

\begin{figure}[htbp]
    \centering
    \subfigure[Traditional RL]
    {
        \includegraphics[width=0.45\linewidth]{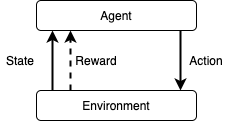}
        \label{fig:trl}
    }
    \subfigure[Multi-Objective RL]
    {
        \includegraphics[width=0.45\linewidth]{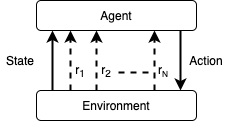}
        \label{fig:morl}
    }

    \caption{Architecture of Reinforcement Learning.}
    \label{fig:rl}
\end{figure}

 MORL is different from tradtional RL in that there are two or more objectives to be optimized simultaneously by the learning agent. \cite{liu2014multiobjective} provides an architecture for a MORL problem, where reward is provided for the learning agent at each step.   Figure \ref{fig:rl} shows the difference between architectures of traditional RL (Figure \ref{fig:trl}) and multi-objective RL (Figure \ref{fig:morl}). In MORL ( Figure \ref{fig:morl}), there are $N$ objectives and $r_i ( 1 \leq i \leq N )$ is the $i^{th}$ reward signal provided by the environment. The architecture illustrates a single agent that has to find an optimal policy for a set of multiple objectives simultaneously. The objectives can be conflicting, as in our case, or they can be independent as well.
 
For each objective i $(1 \leq i \leq N)$ and a stationary policy $\pi$, there is a corresponding state-action value function $Q_{i}^{\pi} $, which satisfies equation \ref{eq:q}.

Let the combined value function for MORL is: 
\begin{displaymath}
MQ^{\pi}(s,a) = \left[ Q_1^{\pi}(s,a), Q_2^{\pi}(s,a), ..., Q_N^{\pi}(s,a)  \right]
\end{displaymath}
where $MQ^{\pi}(s,a)$ is a vector and it also satisfies the Bellman equation (\ref{eq:q}). Then the optimal state-action function will be given as:

\begin{equation}
    MQ^*(s,a) = \max_{\pi} MQ^{\pi}(s,a)
\end{equation}

and the optimal policy $\pi^*$ can be obtained by:

\begin{equation}
    \pi^*(s) = \argmax_a MQ^*(s,a)
\end{equation}

MORL is a combination of multi-objective optimization methods and RL techniques to solve sequential decision making problems with multiple conflicting objectives. We will justify why we formulate smart home power management as a MORL problem in the next section.

\section{Problem Formulation}
\label{section:problem}
 \noindent The case of a smart home power management system is a multi-objective problem with two objectives, viz., minimizing power consumption and maximizing user satisfaction. Ideally, a controller will try to reduce the power consumption as much as it can, given an optimization goal. The trivial solution for the controller will be to turn off all the devices that operate in the smart home. However, this state might not be desirable by the user. Therefore, it is important for a controller to consider user preferences as well. Hence, the goal of minimal power consumption must be achieved by establishing a trade-off with user satisfaction, and not at it's expense.

Based on the importance of an objective function, optimization priorities must be ensured while designing the policies. After appropriately expressing the preferences, we have to design an efficient algorithm that can solve the sequential decision making problems based on observed state transition data. 

\begin{figure}[htbp]
    \centering
    \includegraphics[width=\linewidth]{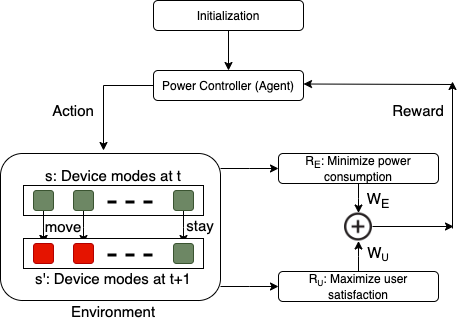}
    \caption{Illustration of Power controller. Detailed explanation of each component is available in the text. }
    \label{fig:arch}
\end{figure}

\subsection{Environment}
\noindent Smart homes usually have smart meters to measure the power consumption for each device operating within it. The power consumption values for every device is independent, and take a fixed number of discrete values. This is because each device operates only in a fixed number of modes and their power consumption in a specific mode remains the same. For example, a simple furnace has only two modes, ON and OFF. In OFF mode, it consumes no power while in ON mode it consumes $x$ (say) units of power. We are assuming that the consumption $x$ remains the same and no degradation of device happens over time, hence causing more energy consumption. Similarly, a washing machine can have three modes of operation, viz, standby, washer and dryer. Let's assume a smart home has $N$ devices. A state in an environment is a vector of energy consumption values (in whatever device modes they are in) of these $N$ devices, as depicted by blocks $s$ and $s'$ in Figure \ref{fig:arch}. 

Let's assume the number of device modes each one operates in is given by a set $D = \{ nD_1, nD_2, ..., nD_N \}$
\begin{displaymath}
Total \ number \ of \ states \ in \ state-space = \prod\limits_{i=1}^{N} nD_i
\end{displaymath}

Increasing the number of devices, or just their modes of operation can lead to state-space explosion. Therefore, in our techniques we choose devices selectively and use data processing (explained in Section \ref{section:exp}) to avoid state-space explosion.

\subsection{Power Controller (Agent)}

\noindent The Agent is a power controller that can change the mode of operation of any of the $N$ devices, consequently changing the energy consumption value. For example, a power controller can turn off the furnace, if it is on, or switch the washing machine to dryer mode from some other mode. However, an agent can also choose not to do anything. Therefore, the Agent can perform either of the two actions, i.e., $MOVE$ or $STAY$, on each device in a state $s$, to move the environment to a state $s'$. For example, in Figure \ref{fig:arch}, the controller changes the mode of operation of device 1 and device 2, from $green$ to $red$ by $MOVE$ action, and chooses to keep the $Nth$ device it its current state by $STAY$ action. 

\subsection{Reward}
\label{section:reward}
\noindent Whenever the proposed agent takes an action on the environment, a reward is calculated on the basis of the state chosen by the controller and the ground truth state from the Smart House Dataset. Since there are two distinct objectives, we formulate the reward functions to incorporate the  power consumption and user satisfaction. Every update of the state-action value function (Equation \ref{eq:q}) is dependent on the reward. Therefore, by integrating the optimizations in reward function, the agent learns the  trade-off between optimization priorities for an optimal state. First we introduce both rewards separately and then we combine them to form a single reward, as shown in Figure \ref{fig:arch}.

\subsubsection{Minimizing Power Consumption}
Let's say the power consumption in the predicted and ground truth state is $P_{s'}$, and $P_s$ respectively. The reward, $R_E$ is given as:

\begin{equation}
\label{eq:re}
	\begin{split}
		R_E = \frac{\sum_d{-(P_{s'} - P_s )}}{D}\\
	\end{split}
\end{equation}

which is the average difference of power consumed by the D devices between predicted state and the desired state. As the agent always tries to maximize the reward, thus we negate the sum in order to achieve state which consumes less power than what the user had chosen. By negating, the state with the least electricity consumption becomes goal state for the power controller.

\subsubsection{Maximizing User Satisfaction}
To model user behavior, we compute the Euclidean distance between the predicted state and the ground truth state. The reward

\begin{equation}
\label{eq:ru}
		R_U = \frac{\sum_d{|{s' - s }|}}{D}\\
\end{equation}

where $d \in D\ devices$.

\subsubsection{Overall reward}
We take a weighted combination of both the rewards, $R_E$ (Equation \ref{eq:re}) and $R_U$ (Equation \ref{eq:ru}), and define overall reward as:

\begin{equation}
\label{reward_eqn}
    R = W_E*R_E + W_U*R_U
\end{equation}
where $W_E$ and $W_U$ are the weights to manipulate the optimization priorities of the two objectives. These weights are treated as hyper parameters during experimentation.

 \subsection{Evaluation}
 
 \noindent The evaluation of power controller's performance is two-fold due to the multi-objective nature of the optimization problem. The reward $R_E$ represents negation of power consumption, therefore, a policy with more positive $R_E$ value is desired. Hence, as we increase the number of iterations, the value of $R_E$ should increase. 
 
 Similarly, the reward $R_U$ represents the likelihood that the next state predicted by the controller, $S_U^{pred}$, matches to the next state that user prefers, $S_U^{real}$. However, to evaluate $R_U$, we introduce a term called ``clash rate'' to get a device level view of clashes. To calculate clash rate:
  \begin{equation}
  \label{eqn:clash}
 	clash \ rate = \sum (S_U^{real} == S_U^{pred})
 \end{equation}
 where ``$==$'' is an element wise comparison that assigns 1 if values do not match and 0 otherwise, and returns an array with 1's and 0's. 
 
 For example, let us say the user wants next state to be $S_U^{real} = [D_1, D_2, D_3, D_3, D_1]$, where $D_is$ represent the device modes at this state. Now, the controller takes an action on the environment to change its state to $S_U^{pred} = [D_2, D_1, D_3, D_1, D_1]$. The clash rate in this case is $3$, as the device modes at index 0, 1, and 3 do not match (assume the vectors are indexed starting from 0). As we increase the iterations to train the power controller more, the clash rate should decrease.

\section{Solutions}  
\label{section:sol}
\noindent MORL approaches can be divided into two groups based on the number of policies to be learned \cite{vamplew2011empirical}:single policy and multiple policy approaches. In our case, the objectives are contrasting, and the availability of data allows us to create a sufficiently good representation of the environment. Therefore, we focus on a single policy approach to solve it.

The aim of single policy approaches is to obtain the best policy which satisfies the optimization priorities as set by the designer, or defined by the application domain. Therefore, based on varying optimization priorities we implemented four variations of a single policy algorithm to find an optimal policy for our two-fold objectives of minimum power consumption and maximum user satisfaction. A single policy approach to solve MORL problems is to formalize an objective function $TQ(s,a)$, which can represent overall preferences in optimization. The approach is very similar to Q-learning with a few modifications, as shown in Algorithm \ref{alg:naive}. The objective function $TQ(s,a)$ is given as the summation of Q-values for all the objectives, and is given as:

\begin{equation}
\label{eqn:tqsa}
    TQ(s,a) = \sum_{i=1}^{N} Q_i(s,a)
\end{equation}

\begin{algorithm}
 \caption{Single Policy Approach to solve MORL}
 \begin{algorithmic}[1]
 \label{alg:naive}

%  \renewcommand{\algorithmicrequire}{\textbf{Input:}}
%  \renewcommand{\algorithmicensure}{\textbf{Output:}}
%  \REQUIRE in
%  \ENSURE  out
%  \textit{Initialisation} :
  \STATE K: the maximum number of episodes
  \STATE N: the number of objectives 
  \STATE Initialize $TQ(s,a)$;
  \STATE Initialize $Q_i(s,a),\ \forall (i<N) $;
  \FOR {each episode $j$ ranging from $1$ to $K$}
  \STATE Fetch $s^j_0, s^j_1$ from samples;
       \STATE Choose $a$ using $TQ(s,a)$ policy using $\epsilon\_greedy$ approach;
       \STATE Take action $a$, $s'$;
       \STATE Compute reward [$r_1$, $r_2$,..., $r_N$ ] based on $s^j_1$, and $s'$;
       \FOR{$i = 1,2,...,N$}
           \STATE $Q_i(s,a) \leftarrow Q_i(s,a) + \alpha[r_i + \gamma \max_{a'} Q_i(s',a') - Q_i(s,a) ]$;
        \ENDFOR
        \STATE Compute $TQ(s,a)$;
        \STATE $s \leftarrow s'$;
%   \IF {($i \ne 0$)}
%   \STATE statement..
%   \ENDIF
  \ENDFOR
 \end{algorithmic} 
 \end{algorithm}
 
 As discussed in Section \ref{section:reward}, we incorporate the optimization priorities using weights $W_E$ and $W_U$ in the reward function. Since $Q(s,a)$ is dependent on the reward function, and $TQ(s,a)$ on $Q(s,a)$, any change in the weight values in equation \ref{reward_eqn} will result in a change of values in $TQ(s,a)$. They are defined as:

\subsection{Single Policy Single Objective}
\noindent As a baseline reference, we implement single policy approach with single objectives. Recall equation \ref{reward_eqn}, the overall reward is defined as the sum of two rewards, one for minimizing power consumption and other for maximizing user satisfaction. Therefore, for single policy with one objective taken at a time has two cases:

\subsubsection{Power Consumption Minimization}
To implement this, we give 100\% optimization priority to power consumption, and set $W_E$ and $W_U$ to $1$ and $0$, respectively in equation \ref{reward_eqn}.

\subsubsection{User Satisfaction Maximization}
To implement this, we give 100\% optimization priority to user satisfaction, and set $W_E$ and $W_U$ to $0$ and $1$, respectively in equation \ref{reward_eqn}.

\subsection{Single Policy Multi Objective}
\noindent The goal of the power controller is to achieve a multi-objective optimization. Therefore, we consider two cases:

\subsubsection{Equal weights}
The case where both objectives are equally as important, and the power controller tries to optimize both. Both $W_E$ and $W_U$ are set to $1$. Based on the policy calculated by the power controller, the action with the maximum summed values is chosen to be executed.

\subsubsection{Weighted Sum}
The weighted sum approach is proven to be effective with multiple objectives in the past. \cite{ngai2011multiple} used it to combine seven vehicle overlapping objectives, and \cite{zeng2010self} used it with a combination of three objectives, viz., degree of the crowd in an elevator, the waiting time, and the number of start-ends. The approach modifies equation \ref{eqn:tqsa} as:

\begin{displaymath}
    TQ(s,a) = \sum_{i=1}^N W_i Q_i (s,a) 
\end{displaymath}

In our case, the $W_is$ are $W_E$ and $W_U$, and we experiment with different values of both to get the best results.

\section{Environment Setup}
\label{section:exp}
\noindent We evaluate the proposed solutions in Section \ref{section:sol} using the Smart* data set for sustainability ~\cite{Barker2012Smart}. As a baseline reference, we consider Q-learning with single objective of power consumption minimization and user satisfaction maximization. We plot the reward and clashes for all four proposed algorithms to contrast the results. In this section, we first briefly explain the data set, then the environment design, and finally the experiments and results.

\subsection{Smart* Data Set for Sustainability}
\label{section:data}
\noindent The data set includes real power consumption readings of multiple devices such as furnace, fridge, washing machine, etc., inside smart homes collected over regular intervals of 30 minutes. Each device has sensors attached to them to record the power consumption after regular time intervals inside seven smart homes\footnote{http://traces.cs.umass.edu/index.php/Smart/Smart}. For our experiments, we used data from smart home A collected over a period of three years. We also use data from smart home B, and C to show transferability of the framework. Note that data from each smart home is similar in nature. The only difference is in the type and the number of devices used to collect data. 

\subsection{Designing the Environment}
\noindent The data set has power consumption values from more than 20 devices for each smart home. We have considered only 5 devices from a smart home: furnace, washing machine, fridge, heater, and kitchen lights. The reasons to do so are:

\begin{itemize}
    \item In a real world scenario, a user does not want the controller to operate on all of the devices in their smart home. 
    \item Formalizing an optimization problem with only a top few devices with maximum power consumption is more realistic and helpful than taking all the possible devices and constraints into consideration.
    \item For simplicity of experimentation.
\end{itemize}

\subsubsection{Data Processing}
In our data set, the power consumption reading for each device took many distinct unique values. For example, the power consumption values for Furnace has $17,000$ unique entries, and that of Fridge is $16,000$. However, a lot of these values are very close and differ only at $4^{th}$/$5^{th}$ decimal place representing a data collection glitch. Since we chose 5 devices, a \textbf{state} in this environment is represented by a vector of size $[D^1_1, ..., D^k_i, ..., D^5_2 ]_{1 \times 5}$ where $D^k_i$ represents the device $D^k$ in mode $i$, and its value is given as the power consumption by device $k$ in mode $D_i$. 

The size of the state space is the cross product of all the unique values taken by each device. Therefore by this convention, if we consider only the furnace and the fridge, the size of state space will be $272$ million ($16,000 \times 17,000$). With such a big state space, the problem becomes very complex to solve, and therefore, to avoid the state space explosion, we cluster the energy consumption values of each device separately to find a fixed \textbf{number of modes of operation} for each device. Intuitively, in real-life, a furnace cannot have 17,000 modes of operation. Therefore, finding device modes with clustering seems to be a fair assumption to make.

\subsubsection{Clustering to assign the modes of operation for each device}

We wanted to find cluster centers of power consumption values for each device individually, which can represent different modes of operation. The modes of operation can be readily available from manufacturer's end but they might not be ideal for our case. For example, suppose a washing machine consumes $x$ power in standby mode, $y$ in wash mode and $z$ in dry mode, and the values, $y$ and $z$ are very close. The manufacturer can say that modes $y$ and $z$ are different, but we have similar readings for the two states, and hence, does not affect our objective. Therefore, we cluster the readings such that each device mode represents a significant amount of change in power consumption from one mode to another. The clustering helps us reduce the state space to a very good extent. 

 \begin{figure}[htbp]
    \centering
    \subfigure[Silhouette analysis for K-means clustering with k = 3 for Duct Heater]
    {
        \includegraphics[width=0.45\linewidth]{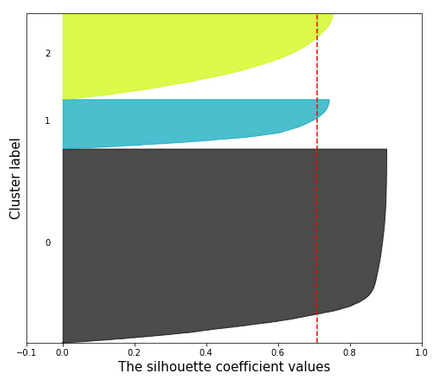}        
        \label{fig:sil}
    }
    \subfigure[Distribution of consumption values of Duct Heater with k = 3]
    {
        \includegraphics[width=0.45\linewidth]{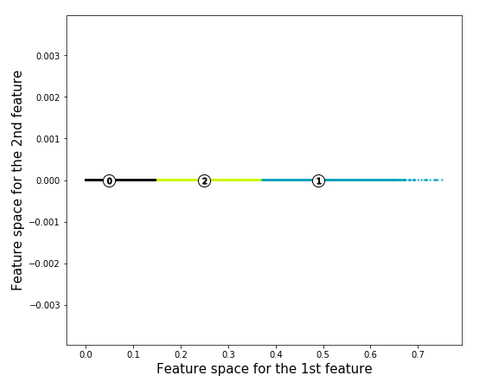}
        \label{fig:dist}
    }

    \caption{Clustering results for Duct Heater from smart home A.}
    \label{fig:clustering}
\end{figure}

First we performed silhouette analysis \cite{ROUSSEEUW198753} to find the optimal number of clusters for each device. We vary $k$, the number of clusters from 2 to 6 assuming it is rare that a device has more than 6 modes of operation. Figure \ref{fig:sil} shows the silhouette plot for various clusters using Duct Heater's electricity consumption data for $k=3$. The clusters are well formed with coefficient value more than the threshold, as can be seen in Figure \ref{fig:dist}. The plot is shown for $k=3$ as it yielded the best silhouette score. Similar experiments are performed with remaining 4 devices.

% \begin{figure}[htb!]
%     \centering
%     \includegraphics[width=0.7\linewidth]{sil.png}
%     \caption{Silhouette analysis for K-means clustering with k = 3 for Duct Heater}
%     \label{fig:sil}
% \end{figure}

% Once we get the optimal number of clusters for Duct Heater, we plot the data with cluster centers. Figure \ref{fig:dist} shows distribution and assigned labels to actual data points.

% \begin{figure}[ht!]
%     \centering
%     \includegraphics[width=0.9\linewidth]{assets/feat.png}
%     \caption{Distribution of consumption values of Duct Heater with k = 3}
%     \label{fig:dist}
% \end{figure}

After clustering, the optimal number of clusters for the chosen devices, Furnace, Washing Machine, Fridge, Duct Heater, and Kitchen Lights is 2, 3, 3, 3, and 5, respectively. Clustering reduces the size of state space from $272$ million+ ($16,000 \times 17,000$) to just 270 ($2 \times 3 \times 3 \times 3 \times 5$)  preventing the state space explosion.

\section{Experiments}
\label{section:newex}
\noindent The first objective is to minimize the total power consumption, and the second objective is to maximize user satisfaction. Our algorithm takes into account 4 hyper-parameters: learning rate ($\alpha$), discount factor ($\gamma$), exploration rate ($\epsilon$), reward prioritization weights $W_E$ and $W_U$. As a baseline, we use Algorithm \ref{alg:naive} with single objectives. Note that if we run the algorithm with a single objective, it becomes the traditional Q-learning algorithm. The clash rate as defined in Equation \ref{eqn:clash} will be maximum for single policy with power consumption minimization objective, and minimum with user satisfaction maximization objective. However, with multi-objectives, the cash rate should be between the two. The overall reward is given as the weighted sum, therefore, reward will be maximum for multi-objective approach. We implemented the solutions discussed in Section \ref{section:sol} as:

\subsection{Single Policy Single Objective}
\noindent The overall reward has two weighted terms, $W_E$ and $W_U$ representing power consumption and user satisfaction, respectively. For the first set of experiments, we focus only on optimizing a single objective by initialising $W_E$ and $W_U$ as $(1,0)$ for power consumption minimization objective, and $(0,1)$ for user satisfaction maximization objective. Hence, in single policy single objective Q-learning formulation, our agent only receives the reward $R_E$ in the former case and the reward $R_U$ in the latter.

We experimented with more than 100 combinations of $\alpha$ and $\gamma$ with $\alpha,\gamma \in (0, 1]$ to find the best hyperparameters. The agent calculates average total reward $R$ and the $clash rate$ for every combination of our hyperparameters over a total of 463 unique states episodes learned over 300 epochs. We decayed the value of $\epsilon$ by a factor of 1.4 every 20 epochs. The set of parameters which gives us the highest average reward and least number of clashes is chosen. The hyperparameters shown in Table \ref{tab:single_obj_params} achieve the best results when our aim is to minimize the average number of clashes to meet each objective individually. 

\begin{table}[htbp]
  \centering
    \caption{Hyper-parameter values for Single Policy Single-Objective Q-learning}
  \begin{adjustbox}{width = \linewidth, center}
    \begin{tabular}{|c|c|c|}
    \hline
    \multicolumn{1}{|p{7.5em}|}{\textit{Objective/Hyper-parameters}} & \multicolumn{1}{p{9.145em}|}{\textbf{Power Consumption Minimization}} & \multicolumn{1}{p{9.145em}|}{\textbf{User Satisfaction Maximization}} \\
    % \midrule

    % \textit{Objective/Hyper-parameters} & \textbf{Power Consumption Minimization} & \textbf{User Satisfaction Maximization} \\
    
    \hline
    \textbf{$\alpha$} & 0.4   & 0.9 \\
    \hline
    \textbf{$\gamma$} & 0.1     & 0.05 \\
    \hline
    \textbf{$\epsilon$} & 0.1     & 0.1 \\
    \hline
    \textbf{$W_E$}    & 1     & 0 \\
    \hline
    \textbf{$W_U$}    & 0     & 1 \\
    \hline
    \end{tabular}%
    \end{adjustbox}

  \label{tab:single_obj_params}%
\end{table}%

\subsection{Single Policy Multi-Objective}

\noindent We divide the experiments for multi-objective approaches into two approaches as discussed in Section \ref{section:sol}: Equal Weights and Weighted-Sum.
As shown in Line 11 of Algorithm \ref{alg:naive}, the update function for the Q-values is different than the normal Q-learning formulation. The equation for the Q-value update is given as:

\begin{equation}
\label{eq:q_naiv}
    Q_{i}(s, a) \: += \: \alpha(r_i + \gamma \max_{a'} Q_i(s',a') - Q_i(s,a)) 
\end{equation}

For equal weights approach, the weights $W_E$ and $W_U$ have been assigned the same value of $1$ representing equal priority for both objectives. For Weighted-Sum approach, we perform experiments by taking approximately $2,300$ combinations of $\alpha$, $\gamma$,$\epsilon$, $W_E$ and $W_U$ with their values within the range (0,1]. The best hyperparameters for the multi-objective approaches are listed in the Table \ref{tab:multi_obj_params}. 

\begin{table}[htbp]
  \centering
    \caption{Hyper-parameter values for Single Policy Multi-Objective Q-learning}
  \begin{adjustbox}{width = \linewidth, center}
    \begin{tabular}{|c|c|c|}
    \hline
    \multicolumn{1}{|p{7.5em}|}{\textit{Approach/Hyper-parameters}} & \multicolumn{1}{p{9.145em}|}{\textbf{Equal Weights Approach}} & \multicolumn{1}{p{9.145em}|}{\textbf{Weighted-Sum Approach}} \\
    % \midrule

    % \textit{Objective/Hyper-parameters} & \textbf{Power Consumption Minimization} & \textbf{User Satisfaction Maximization} \\
    \hline
    \textbf{$\alpha$} & 0.9   & 0.9 \\
    \hline
    \textbf{$\gamma$} & 0.1     & 0.1 \\
    \hline
    \textbf{$\epsilon$} & 0.05     & 0.07 \\
    \hline
    \textbf{$W_E$}    & 1     & 0.3 \\
    \hline
    \textbf{$W_U$}    & 1    & 0.1 \\
    \hline
    \end{tabular}%
    \end{adjustbox}

  \label{tab:multi_obj_params}%
\end{table}%

\section{Results}
\label{section:newres}
\subsection{Average Power}
\label{section:res_rew}
\begin{figure}[htbp]
    \centering
    \includegraphics[width=0.8\linewidth]{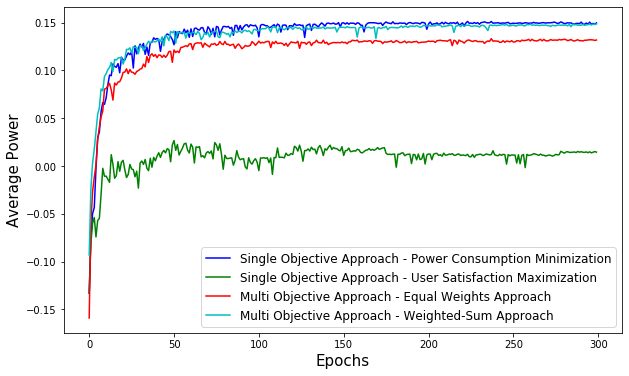}
    \caption{Average Power vs No. of Episodes for different approaches. }
    \label{fig:reward_comparison}
\end{figure}

\noindent To compare the four algorithms proposed to find an optimal policy, we ran them for equal number of epochs using the best hyperparameters obtained for each. Each epoch has $463$ training steps and $450$ validation steps, we plotted the average power for each epoch for comparison. Figure \ref{fig:reward_comparison} shows the average reward for each algorithm.

The average power is maximum for power satisfaction minimization because the policy is getting reward based on the predicted state and user's next state, and it is lowest for user satisfaction maximization due to the fact that if we move from a high power state to a low power state, it will hurt user's satisfaction, which is indeed the desired behavior. The plots for multi-objective approaches always end up between the two single objective ones, representing the trade-off between the two contrasting objectives.

\subsection{Average Number of Clashes}

\begin{figure}[htbp]
    \centering
    \includegraphics[width=0.8\linewidth]{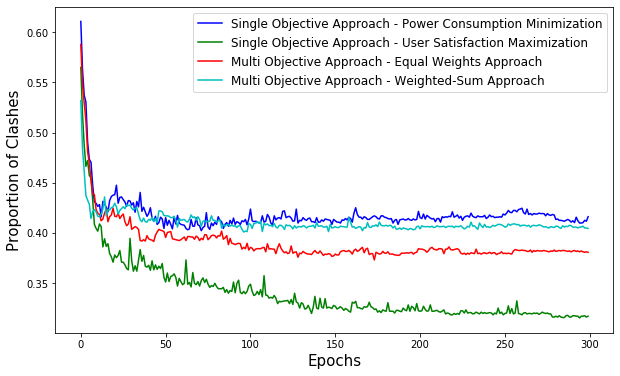}
    \caption{Average Number of Clashes vs No. of Episodes using different approaches. }
    \label{fig:clashes_comparison}
\end{figure}

\noindent Figure \ref{fig:clashes_comparison} shows the combined clash rate for all four algorithms. The experimental parameters are kept same as the previous section. Note that for power consumption minimization the clash rate is highest because no weightage is given to user satisfaction. If we deploy an agent with such policies, the user will get agitated and they will try to override agent's actions, rendering it useless. While on the other hand, an agent with user satisfaction maximization policies will not be helpful in optimizing power consumption. However, if we look at clash rates for multi-objective techniques, they lie between the two single objective approaches, and this clash rate can be adjusted using weights based on user preferences. 

\subsection{Appliance-wise clashes}

\begin{figure}[htbp]
    \centering
    \subfigure[Furnace]
    {
        \includegraphics[width=0.45\linewidth]{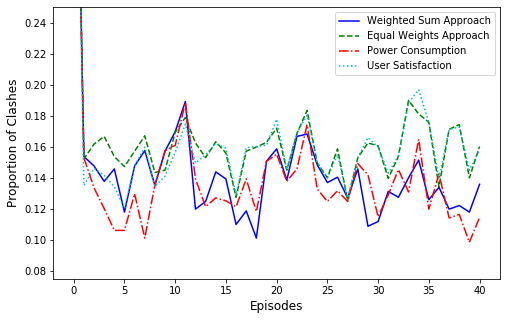}        
        \label{fig:furnace}
    }
    \subfigure[Heater]
    {
        \includegraphics[width=0.45\linewidth]{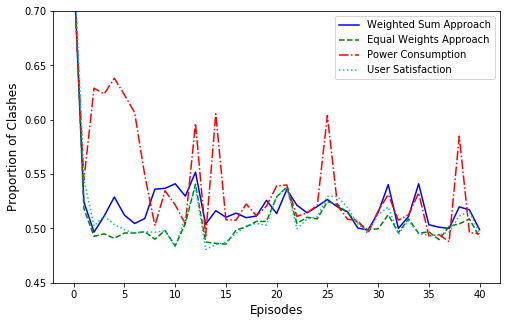}
        \label{fig:heater}
    }
     \subfigure[Lights]
    {
        \includegraphics[width=0.45\linewidth]{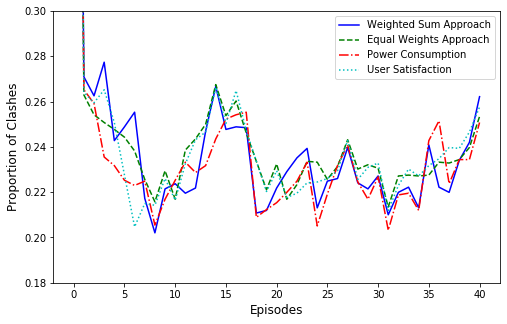}
        \label{fig:lights}
    }
     \subfigure[Refrigerator]
    {
        \includegraphics[width=0.45\linewidth]{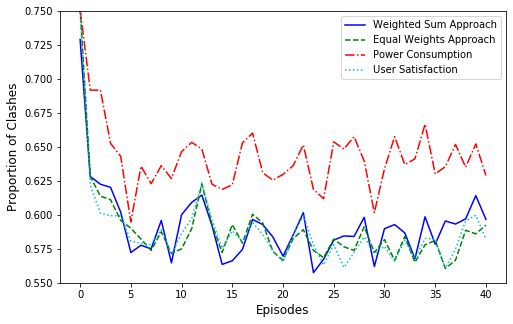}
        \label{fig:fridge}
    }
     \subfigure[Washer]
    {
        \includegraphics[width=0.45\linewidth]{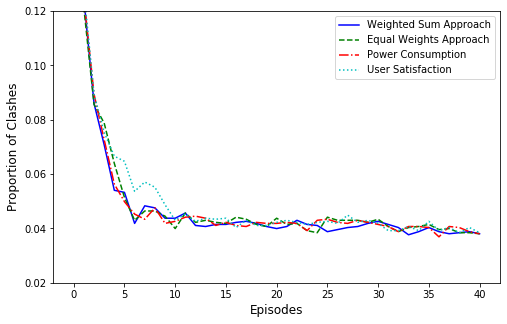}
        \label{fig:washer}
    }

    \caption{Appliance wise clash rate.}
    \label{fig:appliance_wise}
\end{figure}

\noindent We calculate the average number of clashes for each of the five appliances and plot them separately to see the behavior of proposed approaches. Figure \ref{fig:appliance_wise} shows the clash rate for each device. The experimental parameters are kept same as Section \ref{section:res_rew}.

The results for appliances with three device modes is consistent with the overall results with two exceptions of furnace (two device modes) and kitchen lights (5 device modes). 

For furnace, the power consumption minimization approach does not behave as expected. The reason could be the irregular usage and collection of data, as a furnace is used only during colder seasons and the data we used for experiments is collected over a span of three years. For lights, all algorithms fetch nearly same results. The reason can be because lights are used for prolonged times, and there are not many fluctuations in lights' modes of operation. Therefore, the clash rate coincides for user satisfaction, power consumption, and a combination of the two. Hence, Furnace and Lights have very little to contribute to the overall optimization. The results, therefore, suggest that devices that are used regularly and with several fluctuations in device modes at regular intervals should be targeted for optimization.

\subsubsection{Transferability on other smart homes consumption data}
To show that the proposed framework can be applied to power consumption data of multiple smart homes, we choose the best algorithm (weighted sum approach) and run it for smart homes B, and C from the same Smart* data discussed in Section \ref{section:data}.

\begin{figure}[htbp]
    \centering
    \subfigure[Rewards]
    {
\includegraphics[width=0.45\linewidth]{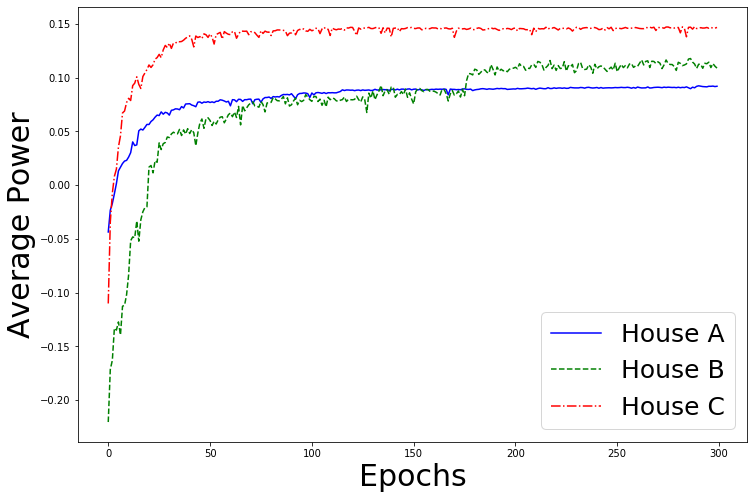}        
        \label{fig:reward_t}
    }
    \subfigure[Clash Rate]
    {
        \includegraphics[width=0.45\linewidth]{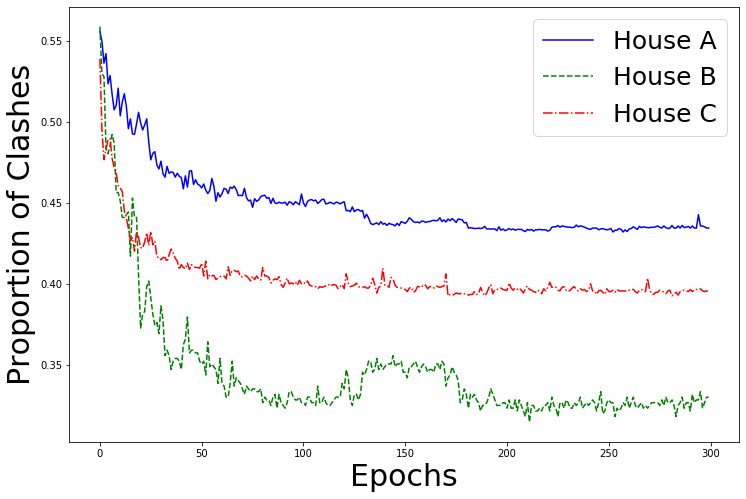}
        \label{fig:clash_t}
    }
    \caption{Average reward and Clashes with Weighted Sum approach smart homes A, B, and C.}
    \label{fig:trans}
\end{figure}

Figure \ref{fig:trans} shows that the rewards increase and the clash rate decreases with increase in number of episodes. Figure \ref{fig:reward_t} and Figure \ref{fig:clash_t} show that the behavior is similar on all three smart homes data.

\section{Conclusion}
\label{section:conclude}
\noindent In this paper, we present a novel multi-objective reinforcement learning technique for power consumption optimization with contrasting objectives of minimizing power consumption and maximizing user satisfaction. We show that both objectives, when considered together, achieve the best optimal policy. Our experimental results show that the proposed multi-objective techniques establish a trade-off between the two objectives. The optimal policy achieves better user satisfaction than power optimization policies and achieves better power consumption than user maximization policies. We show that the devices used regularly in smart homes should be the ones targeted for such optimization purposes. Finally, we also show that the experiments can be performed with other smart home data set to achieve similar results.

\bibliographystyle{IEEEtran}
\bibliography{bare_conf}

\end{document}